\documentclass[10pt,twocolumn,letterpaper]{article}

\usepackage{cvpr}              

\definecolor{cvprblue}{rgb}{0.21,0.49,0.74}
\usepackage[pagebackref,breaklinks,colorlinks,allcolors=cvprblue]{hyperref}

\def\paperID{13539} 
\def\confName{CVPR}
\def\confYear{2025}

\title{DynaDrag: Dynamic Drag-Style Image Editing by Motion Prediction}

\author{Jiacheng Sui\\
Shanghai Jiao Tong University\\
Shanghai, China\\
{\tt\small jcsui01@sjtu.edu.cn}
\and
Yujie Zhou\\
Shanghai Jiao Tong University\\
Shanghai, China\\
{\tt\small yujieouo@sjtu.edu.cn}
\and
Li Niu\\
Shanghai Jiao Tong University\\
Shanghai, China\\
{\tt\small ustcnewly@sjtu.edu.cn}
}

\begin{document}
\maketitle
\begin{abstract}
To achieve pixel-level image manipulation, drag-style image editing which edits images using points or trajectories as conditions is attracting widespread attention. Most previous methods follow move-and-track framework, in which miss tracking and ambiguous tracking are unavoidable challenging  issues. Other methods under different frameworks suffer from various problems like the huge gap between source image and target edited image as well as unreasonable intermediate point which can lead to low editability. To avoid these problems, we propose DynaDrag, the first dragging method under predict-and-move framework. In DynaDrag, Motion Prediction and Motion Supervision are performed iteratively. In each iteration, Motion Prediction first predicts where the handle points should move, and then Motion Supervision drags them accordingly. We also propose to dynamically adjust the valid handle points to further improve the performance. Experiments  on face and human datasets showcase the superiority over previous works.
\end{abstract}    
\section{Introduction}
\label{sec:intro}

\begin{figure*}[!tb]
  \centering
  \includegraphics[width=\textwidth]{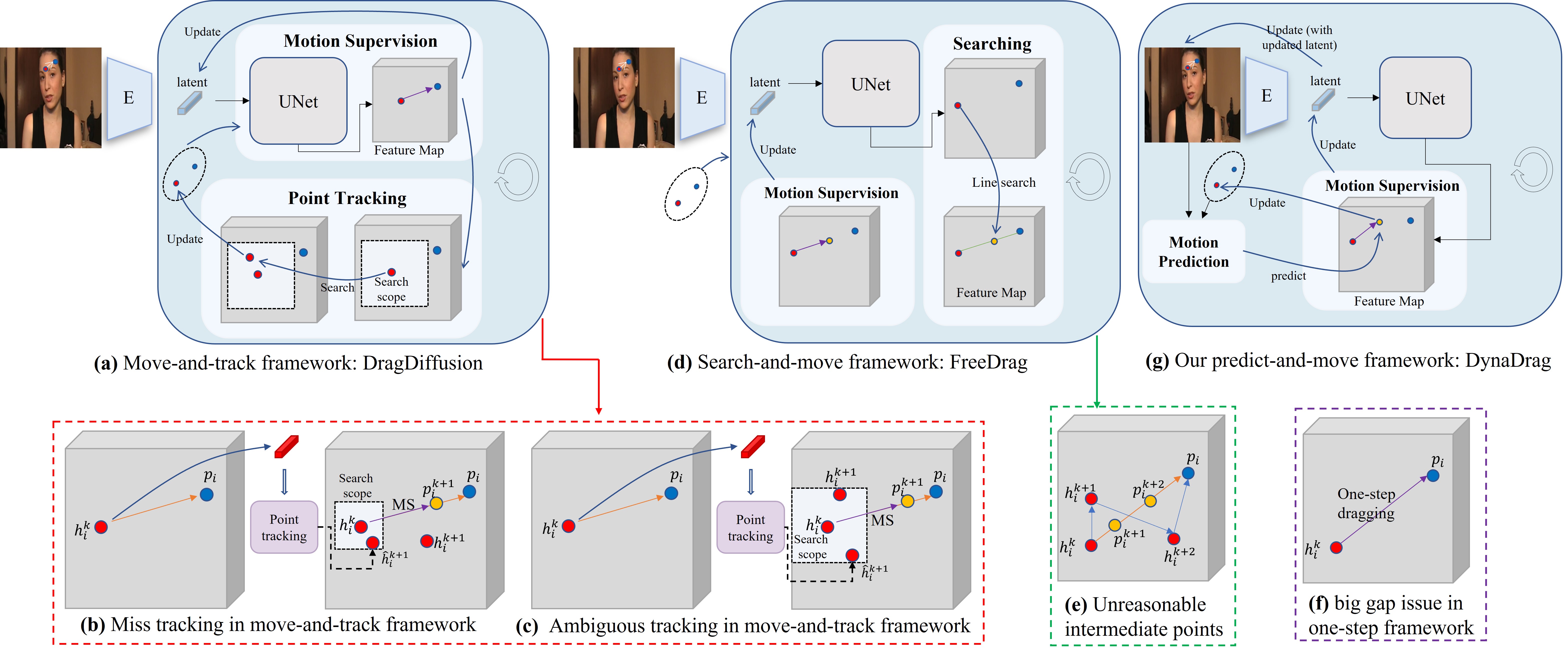}
  \caption{Illustration of existing frameworks, their drawbacks and our framework. Details in these frameworks are omitted. MS in (b,c) represents Motion Supervision. $h_i^k$ is the $i$-th handle point at the $k$-th iteration and $p_i$ represents for its corresponding target point. $p_i^{k+1}$ is the optimization target point at $k$-th Motion Supervision iteration, $\hat{h}_i^{k+1}$ is point tracking searched new handle point whose real location should be $h_i^{k+1}$.
  }
  \label{fig:frameworks_with_drawbacks}
\end{figure*}

Image editing with generative models \cite{mokady2023null,parmar2023zero,roich2022pivotal,hertz2022prompt,brooks2023instructpix2pix,feng2022training} has garnered significant attention due to its diverse range of applications.
 Recently, more and more research has focused on pixel-level manipulation  \cite{endo2022user,pan2023drag,shi2023dragdiffusion,mou2023dragondiffusion,mou2024diffeditor,luo2023readout,nguyen2024edit,hou2024easydrag,liu2024drag,zhang2024gooddrag,adaptivedrag,combing,instantdrag,regiondrag,shi2024instadrag}. Among them, DragGAN \cite{pan2023drag} is one of the earliest works. DragGAN introduces the concept of drag-style image editing, in which users provide an image along with paired handle points and target points. The goal of drag-style image editing is to ``drag" the content at the handle points to the target points. Users can also provide a mask to specify the editable region and the model needs to keep the unmasked region unchanged. DragGAN proposes a move-and-track framework consisting of two iterative steps: motion supervision and point tracking. DragDiffusion \cite{shi2023dragdiffusion} in  \cref{fig:frameworks_with_drawbacks}(a) adopts the same framework as DragGAN, except that DragDiffusion uses the diffusion model \cite{rombach2022high} instead of GAN.

While DragGAN and DragDiffusion have demonstrated remarkable results, they face the challenges such as miss tracking in  \cref{fig:frameworks_with_drawbacks}(b) and ambiguous tracking in  \cref{fig:frameworks_with_drawbacks}(c) in the point tracking step, resulting in degraded image quality. To handle the issues in DragGAN \cite{pan2023drag} and DragDiffusion \cite{shi2023dragdiffusion}, FreeDrag \cite{ling2023freedrag} in  \cref{fig:frameworks_with_drawbacks}(d) proposes a point-tracking-free framework, which is called search-and-move framework in this paper. During experiments, we observe that the enhanced image quality of FreeDrag comes at the expense of compromising the editability and editing precision of the image, which is caused by unreasonable intermediate points in  \cref{fig:frameworks_with_drawbacks}(e). 

For drag-style image editing, there are also one-step editing methods like DragonDiffusion \cite{mou2023dragondiffusion}. Although the one-step approaches can avoid the problems caused by point tracking, they may suffer from poor editability and authenticity due to the large gap between the image to be edited and the target image as illustrated in  \cref{fig:frameworks_with_drawbacks}(f).  Given the issues of one-step dragging, the drag-style editing process should be divided into multiple steps to alleviate the task's degree of difficulty. While point tracking passively determines the new positions of handle points, we proactively predict the positions of intermediate points. We contend that optical flow serves as a reliable indicator of the intermediate point's position. With this insight, we introduce a novel framework for drag-style image editing namely predict-and-move and propose the first method under this framework which is a combination of Motion Prediction (MP) and Motion Supervision (MS). Our framework is illustrated in  \cref{fig:frameworks_with_drawbacks}(g). 

 Specifically, our method performs two iterative steps: Motion Prediction and Motion Supervision. In the Motion Prediction step, we predict the movement of handle points based on the given image along with paired handle points and target points, and then proceed with the Motion Supervision step accordingly. The above framework not only directly avoids the thorny problems brought about by point tracking and improves the rationality of intermediate points, but also makes it possible to dynamically adjust valid handle points. Through Motion Prediction, the movement of handle points is determined, enabling the precise identification of the associated intermediate points. Consequently, we can dynamically auto-select appropriate handle points to achieve better results. In practice, it is evident that not all user-provided handle points have positive contributions,  so we should select reliable handle points for better image editing. 
 
 In general, Motion Prediction in our framework divides and conquers the drag-style image editing task by predicting ``the next movement" of handle points, which avoids annoying problems in point tracking. It can reduce the gap between the image to be edited and the target image, improving the rationality of intermediate points and finally enhancing the editability as well as post-editing authenticity of the image. To summarize, our key contributions are as follows:
\begin{enumerate}
    \item [$\bullet$] We propose a novel predict-and-move framework for drag-style image editing. Under this framework, we propose DynaDrag, which predicts the next movement of handle points and supervises their motion based on the predicted movement.
    \item [$\bullet$] We propose to dynamically select valid handle points, which can help robustly complete the editing task and maintain image quality.
    \item [$\bullet$] Extensive experiments have shown the stability and superiority of DynaDrag in drag-style image editing.
\end{enumerate}

\section{Related Work}
\subsection{Diffusion Models}
Diffusion models are first proposed in \cite{sohl2015deep}. In recent years, with the growth of computing capacity and data scale, DDPM models \cite{ho2020denoising} and DDIM models \cite{song2020denoising} have made it possible to synthesize high-quality images and introduce diffusion models to various tasks \cite{kim2022diffusionclip,xu2023versatile,zhang2023adding,mou2023t2i,karnewar2023holodiffusion}. These models iteratively denoise a random noise to synthesis an image. Latent diffusion models (LDMs) (e.g. Stable Diffusion (SD) \cite{rombach2022high}) are explored to optimize time and memory efficiency..
SD contains an autoencoder and an UNet as denoiser. The autoencoder is responsible for transforming images between the pixel space and the latent space. Both noising and denoising process of LDMs are carried out in the latent space.

\subsection{Drag-style Image Editing}
Existing methods on drag-style image editing can be basically divided into three groups by different frameworks.

\textbf{Move-and-track framework} ( \cref{fig:frameworks_with_drawbacks}(a)): DragGAN \cite{pan2023drag}, DragDiffusion \cite{shi2023dragdiffusion}, StableDrag \cite{cui2024stabledrag}, DragNoise \cite{liu2024drag} EasyDrag \cite{hou2024easydrag}, GoodDrag \cite{zhang2024gooddrag} and AdaptiveDrag \cite{adaptivedrag}. DragGAN introduces a novel approach and methodology for drag-style image editing based on StyleGAN \cite{karras2019style,karras2020analyzing}. DragGAN consists of two alternating steps: motion supervision and point tracking. GANs-based image editing methods are limited by GANs model capacity \cite{goodfellow2014generative,karras2020analyzing}. DragDiffusion \cite{shi2023dragdiffusion} first proposes diffusion-based drag-style image editing method. EasyDrag and StableDrag introduce algorithmic improvements in point tracking while DragNoise and GoodDrag changes optimization target. AdaptiveDrag \cite{adaptivedrag} improves editing performance by clustering semantically similar region.

\textbf{Search-and-move framework} ( \cref{fig:frameworks_with_drawbacks}(d)): FreeDrag \cite{ling2023freedrag} discusses the shortcomings of methods under move-and-track framework which can be summarized as miss tracking and ambiguous tracking. To address these issues, FreeDrag combines line search with backtracking to renew the intermediate targets when dragging and template feature via adaptive updating to improve stability.

\textbf{One-step framework}: DragonDiffusion \cite{mou2023dragondiffusion}, DragAPart \cite{li2024dragapart}, FastDrag \cite{zhao2024fastdrag}, LightningDrag \cite{shi2024instadrag}, LucidDrag \cite{cui2024localize} and InstantDrag \cite{instantdrag}. DragonDiffusion relies on classifier guidance to convert the drag-style image editing manipulation into gradients. FastDrag introduces Latent Warpage Function into this task while LucidDrag leverages an LVLM and an LLM to achieve better results. DragAPart and LightningDrag convert the editing points into conditions, which are then injected into the diffusion process. InstantDrag \cite{instantdrag} trains an optical flow prediction model and utilizes the predicted optical flow as a condition to generate the edited image in one step.

Additionally, DiffUHaul \cite{avrahami2024diffuhaul} is specifically designed for Object Dragging in Drag-Style Image Editing. DiffUHaul mainly introduces modifications and enhancements to the attention mechanism. CLIPDrag \cite{combing} primarily focuses on combining text input with drag signals to resolve ambiguities in the editing process. RegionDrag \cite{regiondrag} introduces region-based editing to enhance the accuracy of drag-style image editing and resolve ambiguities.

In contrast to the aforementioned methods, our proposed predict-and-move framework edits images through multiple steps, proactively predicting the positions of intermediate points. This approach simplifies and rationalizes the editing process.
\section{Methodology}

\label{sec:method}
\subsection{Method Overview}
\begin{figure*}[!tb]
  \centering
  \includegraphics[width=.95\textwidth]{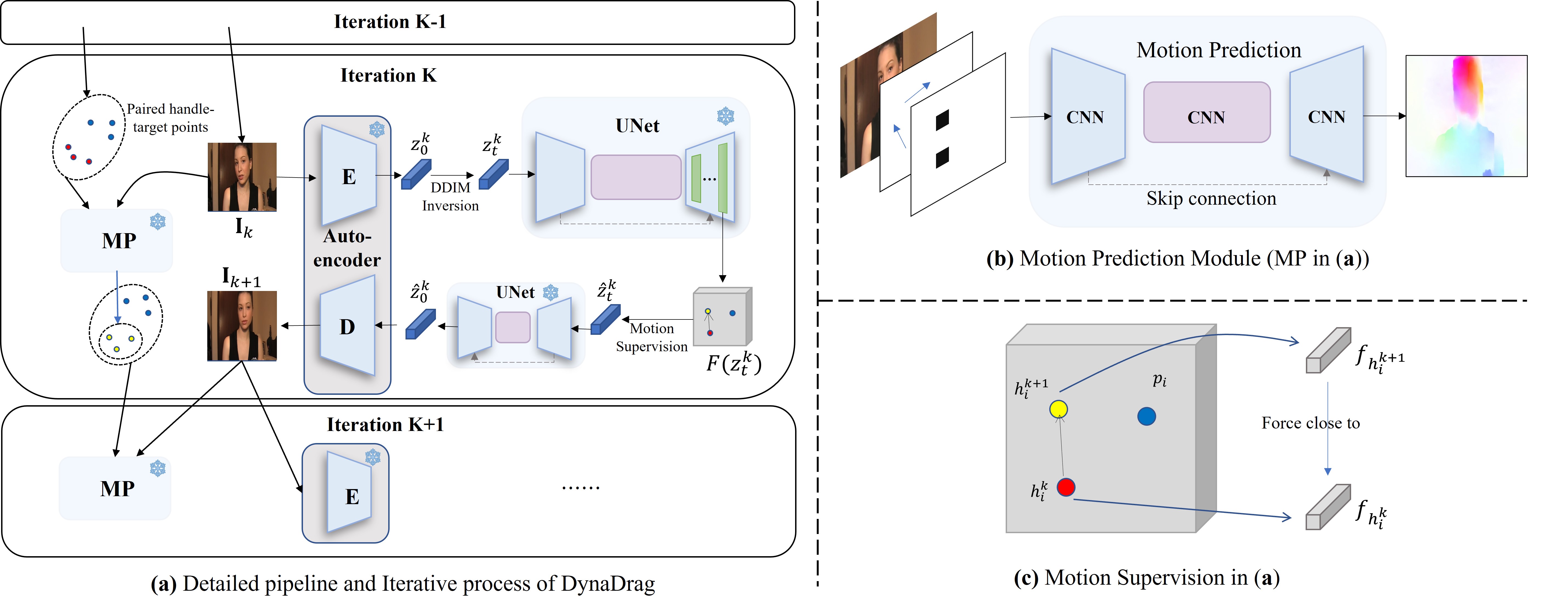}
  \caption{Details and whole pipeline of our method. $h_i^{k}$ in (c) means the $i$-th handle point at the $k$-th iteration, $p_i$ represents for the $i$-th target point while $f_{h_i^{k}}$ is the feature vector at $h_i^{k}$.
  }
  \label{fig:method_and_iteration}
\end{figure*}

Our proposed DynaDrag aims to optimize a specific diffusion latent for drag-style image editing where users select the handle points that will be dragged to their respective target points. To achieve this goal, we first fine-tune a LoRA \cite{hu2021lora} on a diffusion model to reconstruct the input image to better preserve the original image content during the image editing process as suggested in \cite{shi2023dragdiffusion}. After LoRA fine-tuning, we use DDIM Inversion \cite{song2020denoising} to add noise on the input image to obtain a diffusion latent at certain time step $t$. Next, to optimize the $t$-th step diffusion latent, we repeatedly apply Motion Prediction (MP) illustrated in  \cref{fig:method_and_iteration}(b) and  Motion Supervision (MS) illustrated in  \cref{fig:method_and_iteration}(c). The details and whole pipeline of our method is shown in  \cref{fig:method_and_iteration}(a).

For ease of description, we define some notations before introducing our method. We denote the $n$ handle points at the $k$-th predict-and-move iteration as \{$h_{i}^{k} = (x_{i}^{k}, y_{i}^{k})|_{i = 1}^{n}$\} where $\{h_{i}^{0}|_{i = 1}^{n}\}$ are the handle points given by the user and their corresponding target points as \{$p_{i}=(x_{i}, y_{i}) |_{i = 1}^n$\}. The user input image is denoted as $I$, since Motion Prediction and Motion Supervision works iteratively, the user input image can also be denoted as $I_{0}$ which means the input image of the first iteration; the latent code of $k$-th iteration image is denoted as $z_{0}^{k}$; the $t$-th step latent code of $z_{0}^{k}$ obtained by DDIM Inversion is denoted as $z_{t}^{k}$; the feature map of the last UNet block given the $t$-th step latent code $z_{t}^{k}$ as input is denoted as $F(z_{t}^{k})$.

\subsection{Motion Prediction}
\label{sec:motion-prediction}
In this section, we will introduce the details of Motion Prediction and how we dynamically select valid handle points.

\subsubsection{Objective and training details of Motion Prediction.}
We consider the act of dragging handle points to target points as a series of sub-processes, specifically involving the movement of handle points through multiple intermediate points to reach their designated target points.
The purpose of Motion Prediction is to predict the positions of the handle points $h_{i}^{k+1}$ at iteration $k+1$ based on the $k$-th iteration image $I_{k}$ ($I_{0}$ is given by the user), the $k$-th iteration handle points $h_{i}^{k}$, and their target points $p_{i}$.

Specifically, we concatenate a three-channel rgb image $I_{k} \in \mathcal{R}^{3\times H\times W}$, a two-channel delta map $D_{k}\in \mathcal{R}^{2 \times H \times W}$ indicating the distance between the handle points and their corresponding target points, and a single-channel heatmap $G_{k} \in \mathcal{R}^{1 \times H \times W}$ indicating the handle points together as the input of Motion Prediction module. In the delta map $D_{k}$, the value represents the distance ($\Delta x = x_{i}^{k} - x_{i}, \Delta y = y_{i}^{k} - y_{i}$) between the handle point and its corresponding target point at each handle point $h_{i}^{k}$ and zero at all other positions. Considering that the delta map $D_{k}$ is very sparse, following \cite{ma2017pose,dong2018soft}, we concatenate a single-channel heatmap $G_{k}$ as input, in which the value of points in neighborhood ($r=4$) around each $k$-th iteration handle point is 1 and the rest are 0.

Overall, we end up with a tensor $T_{k} \in \mathcal{R}^{6 \times H \times W}$ with 6 channels as input, 3 of which are rgb images and the other 3 channels are used to indicate the positions of handle points and target points. We use SimVP \cite{gao2022simvp}, a simple but effective method for video prediction task, as the structure of the Motion Prediction module, and make slight modification to adapt it to our task. The output of the Motion Prediction module is a two-channel flow map of the same size as input image $I_{k}$, which indicates the position shift value of each point in $I_{k}$. Next, we will elaborate on how to prepare the training data for the Motion Prediction module.

We construct the dataset in the following steps:  1) we use Unimatch \cite{xu2023unifying} to predict the optical flow $f \in \mathcal{R}^{(S-1) \times H \times W \times 3}$ of the given video $v \in \mathcal{R}^{S \times H \times W \times 3}$. 2) We randomly select a frame $v_{s}$ from the $S$ frames as the starting frame, and then extract the editing region with segmentation or detection methods.  3) We randomly sample $1 \sim 7$ handle points in the editing region of $v_{s}$, with the probability proportional to the magnitude of its optical flow $f_{s}$. 4) According to the optical flow predicted by Unimatch, we randomly select a frame in $v_{s+15} \sim v_{s+55}$ as the end frame $v_{e}$, and calculate the ending positions of the handle points in the end frame $v_{e}$ as the target points. 5) The optical flow map $f_{s}$ will be used as ground-truth when training Motion Prediction module. According to the positions of paired handle points and target points, we derive the corresponding delta map $D_{s}$ and heatmap $G_{s}$. More details of dataset construction can be found in the supplementary.

In general, the training loss of the Motion Prediction module is as follows,
\begin{align}
  \mathcal{L}_{mp} = MSE(f_{s}, MP(v_{s}, D_{s}, G_s)),
  \label{eq:motion_prediction}
\end{align}
where $MP(\cdot)$ represents Motion Prediction module.

Unlike previous methods on drag-style image editing, we totally deprecate point tracking and introduce Motion Prediction into this task, making it possible to dynamically adjust valid handle points.

\subsubsection{Dynamically adjusting valid handle points.}
\label{sec:adjust-valid}
In the real-world application scenarios, not all handle points provided by the user are conducive to drag-style image editing. Some handle points may have minimal or even adverse effects on editing, significantly impacting the authenticity and editability of the image. In this section, we will introduce how to dynamically select valid handle points.

\begin{figure}[tb]
  \centering
  \includegraphics[width=.9\linewidth]{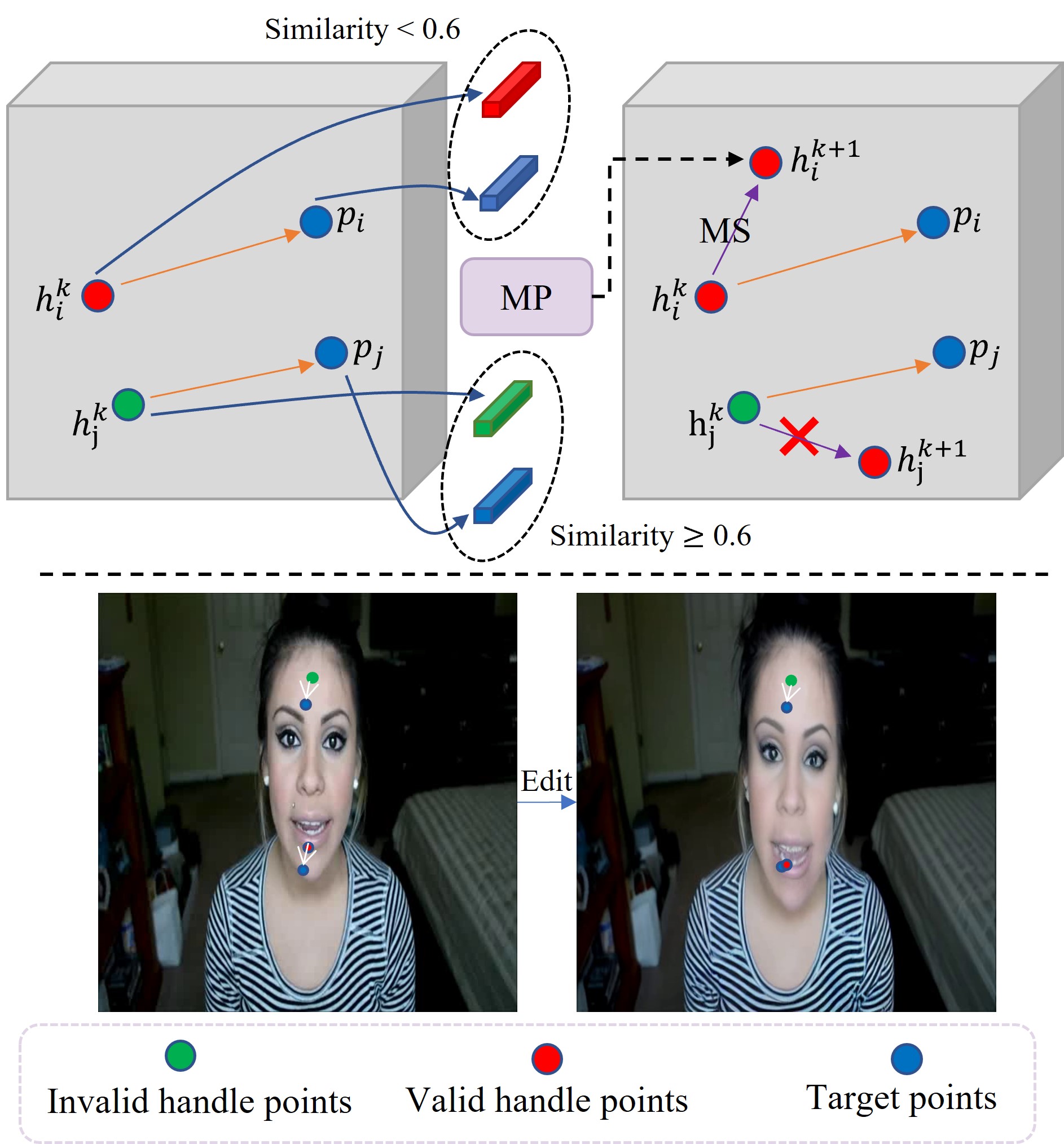}
  \caption{Dynamic Selection strategy. The feature vector of handle point and the feature vector of its corresponding target point are extracted. Cosine similarity between these two feature vector is calculated. Only the pairs with similarity lower than 0.6 are retained. 
  }
  \label{fig:dynamic_selection}
\end{figure}

We find that lower similarity between the feature vectors of handle points and target points in the UNet feature map $F(z_{t}^{k})$ can improve image editability and reduce the chance of generating artificial artifacts, thereby enhancing the authenticity of edited image.
One possible explanation is as follows: high similarity between the feature vectors of handle points and target points suggests minimal disparities in semantics, shape, and other attributes between them. Consequently, the weak supervision signal may mislead the model to infer that adequate adjustment has been done. Conversely, in more challenging scenarios where the similarities between other handle points and target points are low, the model tends to retain the positions of handle points with higher similarity, thereby compromising the authenticity and editability of the image. For example, as shown in the lower sub-figure in  \cref{fig:dynamic_selection}, the lower pair of editing points are intended to edit the lips downward where the semantic gap between the handle point and the target point is large. However, the semantic gap between the upper pair of editing points is relatively small, which makes the network mistakenly think that it has been edited in place. Consequently, as shown in post-edited image, it is evident that the green point almost remains stationary, whereas the red point undergoes adequate and significant editing.

Hence, we filter the paired handle points and target points provided by the user, retaining only those pairs with similarity lower than 0.6 as valid point pairs (as shown in left sub-figure in  \cref{fig:dynamic_selection}). If there is no handle point given by user with similarity less than 0.6, then the handle point with the minimum similarity is used for dragging.

Generally speaking, Motion Prediction avoids the problems caused by point tracking, and reduces the gap between source image and target image by predicting point and making the intermediate points more reasonable. Besides, Motion Prediction supports dynamically adjusting valid handle points.

\subsection{Motion Supervision}
The design of Motion Supervision \cite{pan2023drag,shi2023dragdiffusion} is very simple, and its core is to force the feature vector at the target point in the last feature map $F(z_{t}^{k})$ of the UNet to be close to the feature vector at the handle point. The objective of Motion Supervision is as follows,
\begin{align}
   \mathcal{L}_{\text{ms}}(z^{k}_{t}) &= \sum_{i=1}^{n} \left\|F_{\Omega_{r_{1}}(h^{k+1}_{i})}(z^{k}_{t})-\mathrm{sg}(F_{\Omega_{r_{1}}(h^{k}_{i})}(z^{k}_{t})) \right\|_{1} 
    \nonumber \\
    &\quad +\lambda\left\|(z^{k}_{t-1} - \mathrm{sg}(z^{0}_{t-1}))\odot(1 -M)\right\|_{1},
  \label{eq:motion-supervision}
\end{align}
where $z_{t}^{k}$ is the $t$-th step latent code at $k$-th iteration before Motion Supervision optimization ($z_{t}^{0} = z_{t}$),
$\Omega_{r_{1}}(h^{k}_{i})=\{(x,y) : |x-x^{k}_{i}| \leq r_{1}, |y-y^{k}_{i}| \leq r_{1}\}$ is a neighborhood centered at handle point $h_{i}^{k} = (x^{k}_{i}, y^{k}_{i})$ with a side length of $2r_{1} + 1$, $F_{h_{i}^{k}}(z_{t}^{k})$ is the feature vector at the handle point $h_{i}^{k}$ in the last feature map $F(z_{t}^{k})$ of the UNet, $F_{\Omega_{r_{1}}(h^{k}_{i})}(z_{t}^{k})$ means the feature patch centered at the handle point $h_{i}^{k}$ with side length of $2r_{1} + 1$ in $F(z_{t}^{k})$, $h^{k+1}_{i}$ is the next iteration position of handle point $h^{k}_{i}$ predicted by Motion Prediction module, $M$ is the binary mask given by user to define where can be edited in the image, $sg(\cdot)$ is the gradient stop operator, for example, the backward propagation of $F_{\Omega_{r_{1}}(h^{k}_{i})}(z^{k}_{t})$ will be stopped for the term $\mathrm{sg}(F_{\Omega_{r_{1}}(h^{k}_{i})}(z^{k}_{t}))$. It is worth noting that in the second term of  \cref{eq:motion-supervision} which is optimized based on the user-provided mask, we utilize the diffusion latent code at $t-1$ step for supervision instead of the UNet feature map. Finally, after 5 gradient descent optimization steps, we obtain the diffusion latent code $\hat{z}_{t}^{k}$ for the subsequent predict-and-move iteration:
\begin{align}
    \hat{z}^{k,i+1}_{t} = \hat{z}^{k,i}_{t} - \eta\cdot\frac{\partial \mathcal{L}_{\text{ms}}(\hat{z}^{k,i}_{t})}{\partial \hat{z}^{k,i}_{t}}, i = (0, 1, 2, 3, 4),
\end{align}
where $\hat{z}^{k,0}_{t} = z^{k}_{t}$, $\hat{z}^{k}_{t} = \hat{z}^{k, 5}_{t}$, $\eta$ is learning rate.

After completing Motion Supervision for each iteration $k$, we send the optimized latent code $\hat{z}^{k}_{t}$ to UNet for denoising and obtain the edited image $I_{k+1}$ by the decoder. However, solely employing DDIM to handle the optimized latent code may result in a deterioration of image quality, manifesting as an increase in artificial artifacts that compromise the authenticity of the image. We hypothesize that this phenomenon could be attributed to the destruction of a segment of the original image's latent code throughout the optimization process. To minimize the impact of this problem, according to \cite{mou2023dragondiffusion,shi2023dragdiffusion}, replacing KV seems useful to maintain consistency between the pre-edited and post-edited images. In particular, we guide the denoising process of $\hat{z}^{k}_{t}$ with the denoising process of $z_{t}^{0}$ to preserve the information in the original image. As suggested by \cite{mou2023dragondiffusion,shi2023dragdiffusion}, in the inference stage of UNet denoising, we replace the key and value in self-attention modules obtained from $\hat{z}^{k}_{t}$ with the corresponding parts of $z_{t}^{0}$. 

\subsection{Iterative process of Motion Prediction and Motion Supervision}

Motion Prediction and Motion Supervision are iteratively performed as shown in  \cref{fig:method_and_iteration}(a). When the user provides an image $I_{0}$ along with its paired handle points $h_{i}^{0}$ and target points $p_{i}$, the predicted flow map $f^{pred}$ is first obtained through the Motion Prediction module, and the corresponding positions of the handle points in the next iteration step are calculated as:
\begin{align}
    h^{k+1}_{i} = h^{k}_{i} \oplus f^{pred}_{h^{k}_{i}}, i = 0, 1 \cdots n,
\end{align}
where $f^{pred}_{h^{k}_{i}}$ is the 2-channel vector ($\Delta x, \Delta y$) of pixel position $h^{k}_{i}$ in flow map $f^{pred}$, $\oplus$ means bit-wise addition.

Image $I_{0}$ is encoded as $z_0^0$ through a encoder and noised as $z_{t}^{0}$ through DDIM Inversion.
Next, Motion Supervision optimizes diffusion latent code $z_{t}^{0}$ to ``move" handle points $h_{i}^{0}$ towards $h_{i}^{1}$ predicted by Motion Prediction module and then obtain $\hat{z}_{t}^{0}$. After that, we denoise $\hat{z}_{t}^{0}$ to $\hat{z}_{0}^{0}$ and  input $\hat{z}_{0}^{0}$ into the decoder of autoencoder to obtain the image $I_{1}$ after first editing. It is worth mentioning that during the denoising stage of $\hat{z}_{t}^{0}$ to $\hat{z}_{0}^{0}$, replacing KV strategy is activated. Now, $I_{1}$ and $h_{i}^{1}$, have replaced the positions of $I_{0}$ and $h_{i}^{0}$, and will proceed to the next iteration of Motion Prediction and Motion Supervision loops. The above procedure continues, until the handle points are close enough to the target points or the maximum iteration is reached.

\begin{figure}[!tb]
  \centering
  \includegraphics[width=\linewidth]{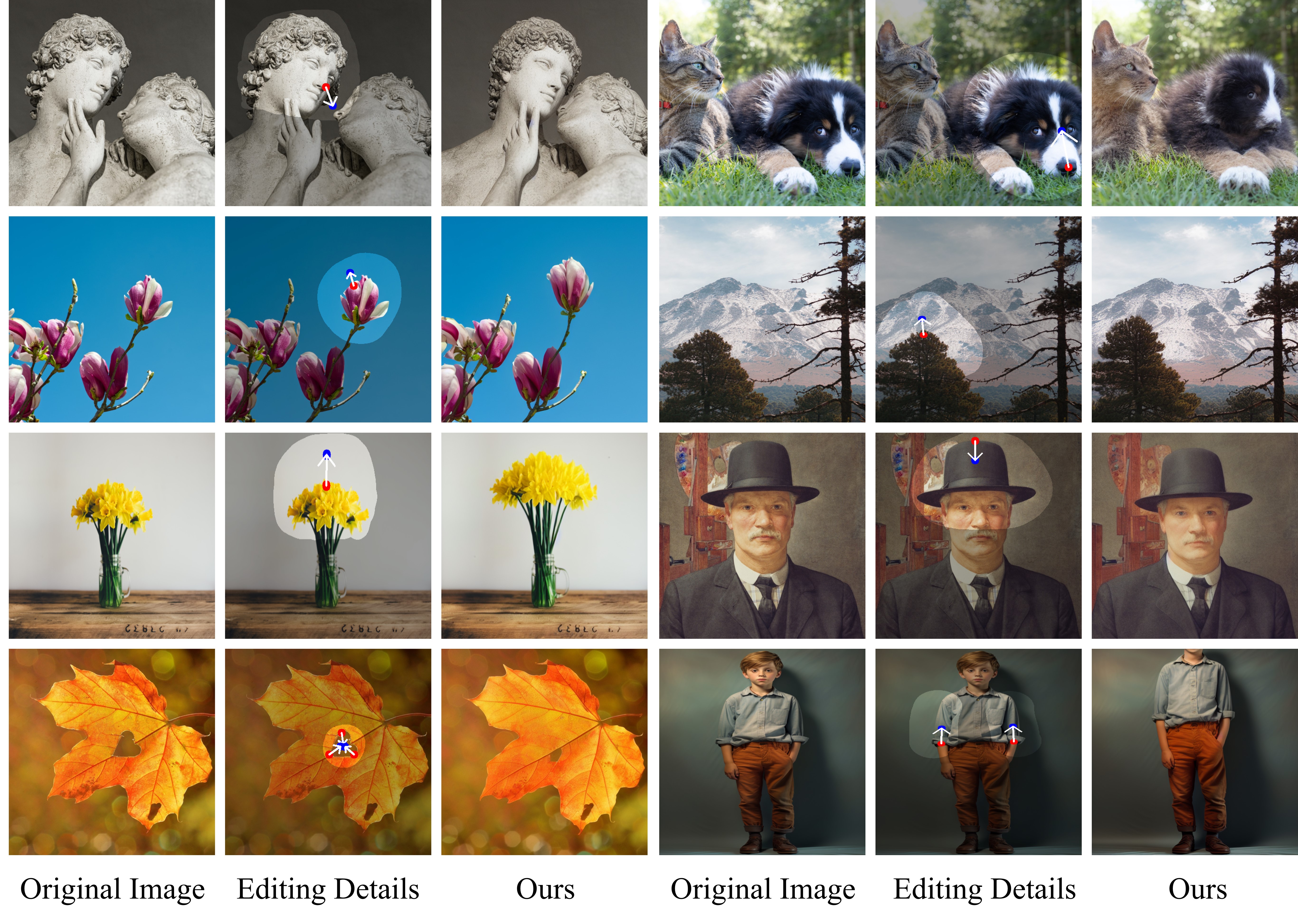}
  \caption{Generalization Study on Motion Prediction. Motion Prediction trained on FaceForensics++ dataset \cite{rossler2019faceforensics++} is used to test on DragBench.
  }
  \label{fig:generalization}
\end{figure}
\section{Experiment}
\label{sec:experiments}

\subsection{Datasets}

\begin{table*}[!t]
  \centering
  \begin{tabular}{@{}c|cccc|cccc@{}}
    \toprule
    Dataset & \multicolumn{4}{c|}{FaceForensics++} & \multicolumn{4}{c}{Ted-talks}\\
    Metric & FID$\downarrow$ & MSE$\downarrow$ & LPIPS$\downarrow$ & CLIP SIM$\uparrow$ & FID$\downarrow$ & MSE$\downarrow$ & LPIPS$\downarrow$ & CLIP SIM$\uparrow$ \\
     &  & ($\times 10^{3}$) &  &  &  &  ($\times 10^{3}$) &  &   \\
    \midrule
    DragDiff \cite{shi2023dragdiffusion} & \underline{51.37} & 1.304 & 0.1564 & 0.9133 & 91.77 & 2.377 & 0.3908 & 0.8254 \\
    DragDiff $+$PIPS2\cite{zheng2023pointodyssey} & 55.93 & 1.617  & 0.1913 & 0.9134 & 69.55 & 1.785 & 0.3534 & 0.8682 \\
    DragonDiffusion \cite{mou2023dragondiffusion} & 79.22 & 1.575  & 0.2045 & 0.8332 & 87.46 & 1.350 & 0.3176 & 0.7777 \\
    FreeDrag \cite{ling2023freedrag}& 52.27 & 1.470  & 0.1674 & 0.9160 & \underline{51.80} & 1.144 & \underline{0.2577} & \underline{0.9181} \\
    EasyDrag \cite{hou2024easydrag} & 59.15 & 1.232  & 0.1675 & 0.8948 & 63.81  & 1.135 & 0.2834  & 0.8579 \\
    DragNoise \cite{liu2024drag}& 58.61 & {\bf 1.124}  & 0.1685 & 0.9106 & 56.33 & {\bf 0.922} & {\bf 0.2433} & 0.8997 \\
    InstantDrag \cite{instantdrag}& 56.48 & 1.382  & 0.1696 & {\bf 0.9221} & 64.35 & 1.707 & 0.3215  & 0.9022  \\
    LightningDrag \cite{shi2024instadrag}&57.00  & 1.244 & \underline{0.1496} & 0.8930 & 77.38 & \underline{1.107} & 0.3031 & 0.8082 \\ 
    Ours & {\bf 48.04} & \underline{1.147} & {\bf 0.1397 }& \underline{0.9205} & {\bf 51.51} & 1.224 & 0.2649 & {\bf 0.9223} \\
  \bottomrule
  \end{tabular}
  \caption{Quantitative evaluation on FaceForensics++ dataset and Ted-talks dataset. Lower FID score suggests better image fidelity, while lower MSE, lower LPIPS and higher CLIP SIM score indicates more precise image manipulation. 
  }
  \label{tab:baseline}
\end{table*}

In this paper, we conduct experiments on face dataset FaceForensics++ \cite{rossler2019faceforensics++} and human dataset Ted-talks \cite{siarohin2021motion}. There are two main reasons for choosing these datasets. Theoretically, the variations in human bodies and faces are the most diverse. Compared to natural scenes like buildings, trees, and natural objects like apples, human bodies and faces are flexible, exhibiting not only positional and postural changes but also variations in size and shape. Therefore, comparing methods on facial and human body datasets is the most effective way to assess their editing capabilities. From an application perspective, users have the highest demand for editing the size, shape, and posture of faces and human bodies, and there are clear industrial application scenarios for this.

To assess the generalization performance of Motion Prediction, we also conduct experiments on DragBench \cite{shi2023dragdiffusion}. Detailed generalization study can be found in \cref{sec:generalization}.

\subsection{Implementation Details}

In all our experiments, we adopt SimVP \cite{gao2022simvp} as the Motion Prediction model and make slight modification to adapt it to our task (detailed configuration and modifications of SimVP are illustrated in Supplementary). We use Stable Diffusion 1.5 \cite{rombach2022high} as our diffusion model for Motion Supervision. 

Before editing image, we first train a LoRA \cite{hu2021lora} and inject it into all attention modules in the denoiser of Stable Diffusion. We set the rank of LoRA to 16 and fine-tune the LoRA using AdamW \cite{loshchilov2017decoupled} optimizer with a learning rate of $2 \time 10^{-4}$ for 200 steps. During Motion Supervision when the latent code is optimized, we schedule 50 steps for DDIM \cite{song2020denoising} and optimize the 50-th step latent code (the first step latent code when denoising). We set the prompt to an empty string, and use Adam \cite{adam} optimizer with a learning rate of 0.01. The maximum iteration of Motion Prediction and Motion Supervision is set to be 25. The hyperparameter $r_{1}$ in \cref{eq:motion-supervision} is set to be 1. $\lambda$ in \cref{eq:motion-supervision} is an user-defined parameter with default value of 0.1. Users may increase $\lambda$ to better preserve the content of unmasked region.

\subsection{Qualitative Evaluation}

\begin{figure*}[tb]
  \centering
  \includegraphics[width=.9\textwidth]{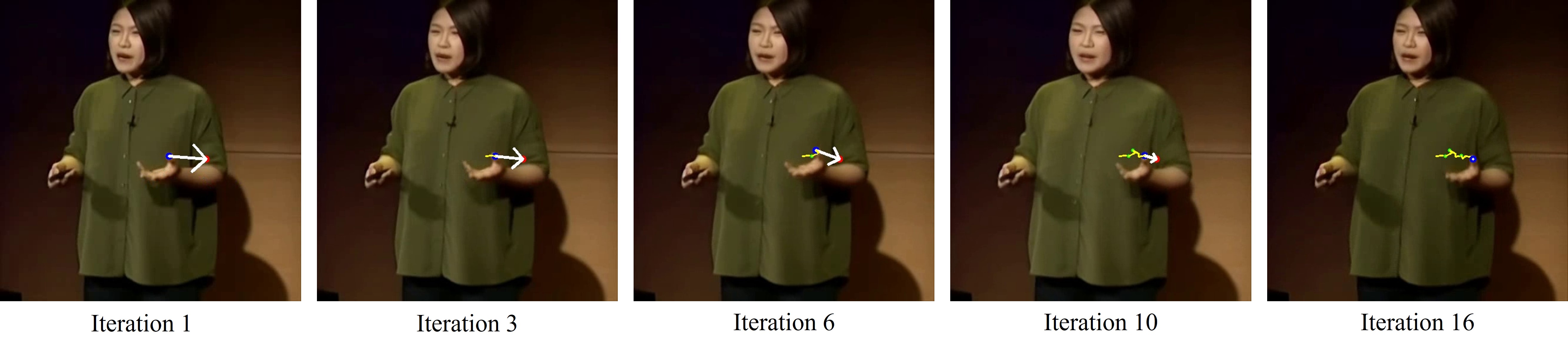}
  \caption{In each iteration, the blue points represent the handle points, the red points denote the target points, the green points indicate the intermediate points in some of the previous iteration steps (to better illustrate the trajectory, we did not plot all the intermediate points), and the yellow line represents the trajectory formed by the intermediate points. It can be observed that the trajectory of the intermediate points does not form a straight line, as would be the case in a typical move-and-track framework. Instead, Motion Prediction allows the handle points to move towards the target points in a smoother and more natural manner (the trajectory generated by Motion Prediction more closely aligns with the actual video trajectory, thereby reducing the difficulty of latent code editing. As a result, the transition between latent codes becomes more natural and smooth).
  }
  \label{fig:effectiveness-analysis}
\end{figure*}

\begin{table}[!tb]
  \centering
  \begin{tabular}{@{}ccccc@{}}
    \toprule
    Dataset & \multicolumn{4}{c}{FaceForensics++} \\
    Metric & FID$\downarrow$ & MSE$\downarrow$ & LPIPS$\downarrow$ & CLIP SIM$\uparrow$ \\
      &  & ($\times 10^{3}$) &  &    \\
    \midrule
    Ours w/o DS & 51.85 & 1.231 & 0.1529 & 0.9152 \\
    Ours w/ RS & 49.30 & 1.171 & 0.1426 & 0.9180 \\
    Ours w/ FDS & 49.41 & {\bf 1.136}  & 0.1400 & 0.9204 \\
    Ours w/ ADS & {\bf 48.04} & 1.147  & {\bf 0.1397} & {\bf 0.9205} \\
  \bottomrule
  \end{tabular}
  \caption{Ablation study on Dynamic Selection (DS). 
  }
  \label{tab:ablation}
\end{table}

\begin{figure*}[tb]
  \centering
  \includegraphics[width=.95\textwidth]{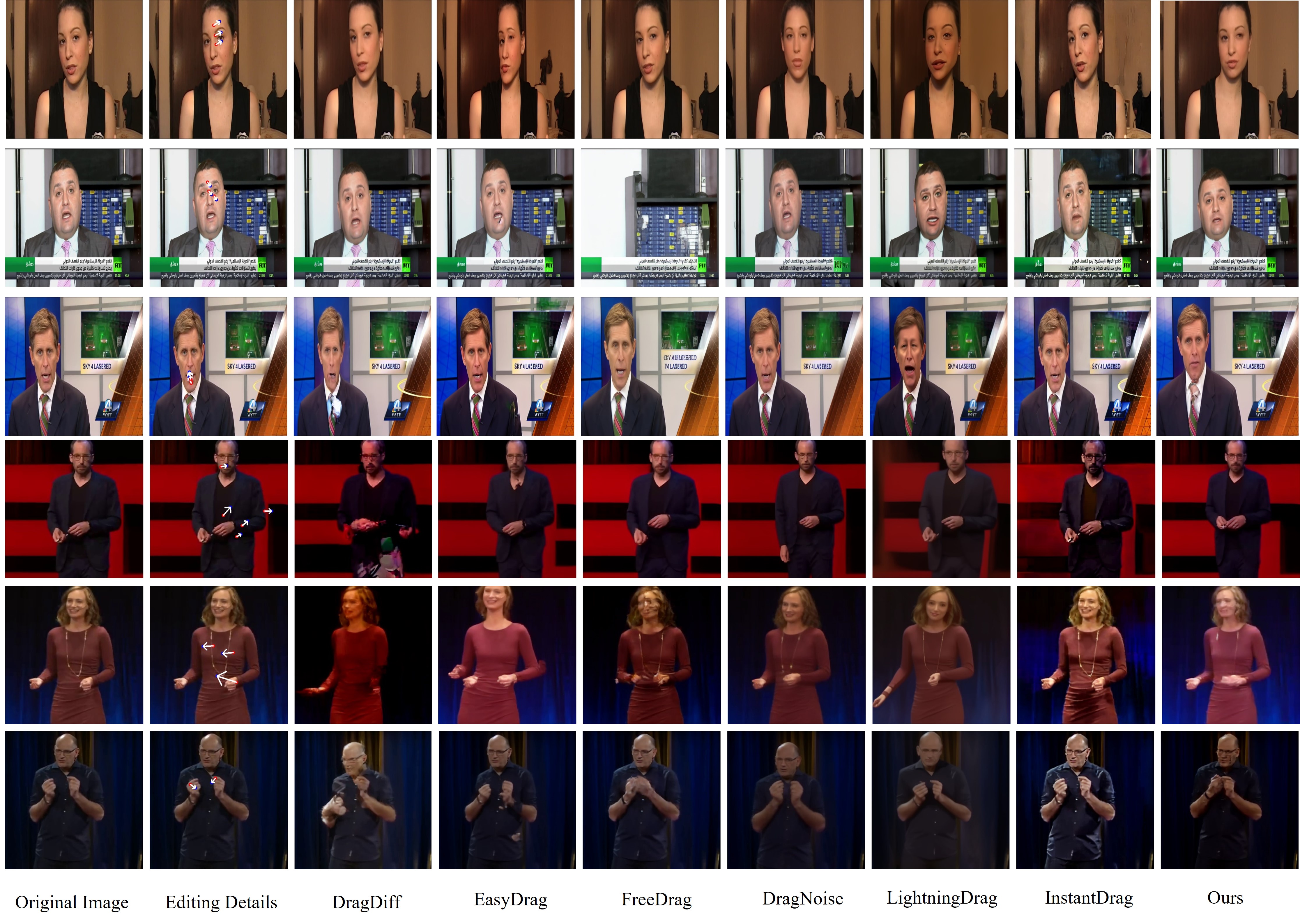}
  \caption{Qualitative evaluation on FaceForensics++ dataset (upper sub-figures) and Ted-talks dataset (lower sub-figures). Our proposed DynaDrag outperforms baseline methods in terms of editing accuracy and image fidelity.
  }
  \label{fig:qualitative-evaluation}
\end{figure*}

As depicted in  \cref{fig:qualitative-evaluation}, our proposed DynaDrag demonstrates superior editability and editing accuracy, enabling precise manipulation of handle points to their corresponding target points to achieve specific editing goals (e.g. reorienting the head, adjusting facial expressions in the FaceForensics++ dataset, and altering hand positions in the Ted-talks dataset). Furthermore, our approach effectively mitigates artifacts that could significantly degrade image fidelity. It is noteworthy that one-step methods, in contrast to methods under move-and-track framework like DragDiffusion, DragDiffusion + PIPS2 (which replaces point tracking in DragDiffusion with PIPS2 tracking \cite{zheng2023pointodyssey}), FreeDrag, and our method, often lead to severe artifacts, while methods under move-and-track framework encounter issues with miss tracking and ambiguous tracking, resulting in undesired editing. Overall, the editing outcomes demonstrate the superiority of our method in terms of editing accuracy and image fidelity over previous methods.

\subsection{Quantitative Evaluation}

This section presents evaluations conducted to assess the performance of our method and existing methods. We compare our method against eight baselines—DragDiffusion \cite{shi2023dragdiffusion}, DragDiffusion + PIPS2, and DragonDiffusion \cite{mou2023dragondiffusion}, FreeDrag \cite{ling2023freedrag}, EasyDrag \cite{hou2024easydrag}, DragNoise \cite{liu2024drag}, InstantDrag \cite{instantdrag}, LightningDrag \cite{shi2024instadrag}—on the FaceForensics++ \cite{rossler2019faceforensics++} and Ted-talks \cite{siarohin2021motion} datasets. DragGAN \cite{pan2023drag} is based on StyleGAN \cite{karras2019style,karras2020analyzing}, but no pre-trained checkpoints are available for these datasets. Specifically, FID \cite{heusel2017gans} is used to evaluate the image quality of edited image. MSE, LPIPS \cite{zhang2018unreasonable} and CLIP similarity \cite{radford2021learning} between edited image and target image are used to evaluate the editing accurary.

As shown in \cref{tab:baseline}, our proposed DynaDrag attains better performance on most metrics testing on FaceForensics++ dataset and Ted-talks dataset which means it outperforms previous baselines in terms of precise pixel-level drag-style image editing and image fidelity.

\subsection{Ablation Study}

In the section, we conduct ablation experiments on the effectiveness of Dynamic Selection (DS) on handle points. Dynamic Selection can be divided into two types based on their level of participation. First Dynamic Selection (FDS) means selecting the valid handle points based on feature vector similarity ONLY in the first iteration when editing and keeping the selected handle points valid for the rest iterations while other handle points are abandoned. All Dynamic Selection (ADS) means applying Dynamic Selection to all iteration when editing which signifies that the valid handle points can be different in different iterations.

As shown in  \cref{tab:ablation}, Dynamic Selection is really helpful to improve the editablity and quality of the image. It should be noted that DS will reduce the number of valid handle points when dragging to edit images. The experiment in DragGAN \cite{pan2023drag} indicates that increasing handle points may be detrimental to maintaining the authenticity of the image. To exclude the influence of the number of handle points and further prove the improvement Dynamic Selection bring about, we conduct experiment of Randomly Selecting (RS) the same number of valid handle points as ADS based on our method whose result is shown in the second line in  \cref{tab:ablation}. FDS and ADS beats RS in almost all metrics testing on FaceForensics++ dataset, which validates the superiority of DS.

\subsection{Effectiveness Analysis of DynaDrag}
\label{sec:effectiveness}

This section analyzes why DynaDrag outperforms the baseline methods. In \cref{fig:effectiveness-analysis}, we present the iterative editing process on one of the frames from FaceForensics++ dataset \cite{rossler2019faceforensics++}. As shown in \cref{fig:effectiveness-analysis}, the image depicts a speaker whose gestures and hand movements enhance the liveliness of the presentation. Unlike move-and-track methods, which place intermediate points along a straight line between handle points and target points, DynaDrag predicts the next intermediate point for the handle point before moving it. This approach results in more reasonable and natural placements of intermediate points, as well as a smoother trajectory for the handle point movement. The well-predicted intermediate points and their natural trajectories also facilitate the handle point's movement towards the expected intermediate point in each iteration, eventually reaching the target point.

\subsection{Generalization Study}
\label{sec:generalization}

We conduct generalization experiments on DragBench \cite{shi2023dragdiffusion} which is a benchmark dataset for Drag-Style Image Editing tasks that includes a wide range of scenarios. DragBench features various scenes, such as humans, animals, and natural landscapes. We study our method's generalization ability by directly employing a Motion Prediction model trained on FaceForensics++ dataset \cite{rossler2019faceforensics++}. Experiments showcase that Motion Prediction demonstrates strong generalization capabilities. For instance, as depicted in  \cref{fig:generalization}, despite being trained on the FaceForensics++ dataset and encountering a notable gap between the FaceForensics++ dataset and DragBench, Motion Prediction remains effective in many scenarios.
\section{Conclusion}

In this paper, we propose DynaDrag, a drag-style image editing method with completely new framework of predict-and-move, making it possible to dynamically adjust the valid handle points when editing the given image. Comprehensive quantitative evaluation, qualitative evaluation, and ablation experiments showcase the impressive performance and superiority of our proposed method. 

\clearpage
\setcounter{page}{1}
\maketitlesupplementary

\noindent In this Supplementary, we provide details regarding the architecture of Motion Prediction in  \cref{sec:mp_arch}, analyse the editing time of our method in  \cref{sec:editing-time}, describe the details of dataset construction in  \cref{sec:dataset_construction}, present additional qualitative results in  \cref{sec:more-quality}, show the visual cases of ablation study on dynamic selection in  \cref{sec:visual-ablation} and discuss the limitations of our method in  \cref{sec:limitation}.

\section{Architecture of Motion Prediction}
\label{sec:mp_arch}

As shown in \cref{fig:method_and_iteration}, we use SimVP \cite{gao2022simvp}, a simple but effective method for video prediction task, as the base model of our Motion Prediction module and make slight modification to adapt it to our task.

Given its focus on video prediction tasks, the design of SimVP must account for both spatial relationships within images and temporal relationships across consecutive frames. Comprising an encoder, translator, and decoder, all of which have CNN structures, SimVP serves distinct purposes: the encoder extracts spatial features, the translator captures temporal evolution, and the decoder combines spatio-temporal details to predict future frames. Specifically, when provided with $T$ adjacent frames in an input shape of $(B, T, C, H, W)$, the encoder reshapes this input to a tensor shape of $(B \times T, C, H, W)$ and applies convolution with $C$ channels on $(H, W)$ to extract spatial information within each frame, yielding a tensor shape of $(B \times T, C_e, H_e, W_e)$. The translator then reshapes the encoder's output to a tensor shape of $(B, T \times C_e, H_e, W_e)$ and convolves $T \times C_e$ channels on $(H_e, W_e)$ to capture temporal relationships between frames. Finally, the decoder integrates temporal information from the translator, merges spatial information from the encoder via skip connections, and ultimately generates the predicted video frame.

In our task, we predict the motion direction of each pixel in the image based on the provided image, handle points, and target points. As temporal features are unnecessary, our focus lies solely on extracting spatial features. To achieve this, we directly concatenate the 3-channel RGB image, delta map, and heatmap into a 6-channel tensor shape of $(B, 6, H, W)$ for input. We eliminate the reshaping step in the translator and maintain the skip connection between the encoder and decoder. Additionally, we modify the output channel count of the decoder from 3 to 2, indicating the movement direction of each pixel in the image. The other default configurations of SimVP, such as the number of hidden channels, remain unchanged in our task.

\section{Editing Time Analysis}
\label{sec:editing-time}
To investigate the editing time of our method, we conducted tests on 500 images with a resolution of $512 \times 512$ per dataset using an NVIDIA RTX 3090 GPU. Average editing time is calculated. As shown in  \cref{tab:time}, the LoRA finetuning process requires approximately 57 seconds per image. The editing phase, comprising Motion Prediction and Motion Supervision, takes 406.8 seconds per image for the FaceForensics++ dataset \cite{rossler2019faceforensics++} and 326.6 seconds for the Ted-talks dataset \cite{siarohin2021motion}. As Motion Prediction is performed in pixel space, DDIM \cite{song2020denoising} sampling is necessary after each Motion Supervision step. Our experiments reveal that DDIM sampling accounts for the majority of the editing time, whereas the optimization process in the Motion Supervision step is relatively faster.

\begin{table*}[tb]
  
  \centering
  \begin{tabular}{@{}c|cccc|cccc@{}}
    \toprule
    Fine-tune LoRA & \multicolumn{8}{c}{Editing phase} \\
    \makebox[.2\textwidth][c]{}& \multicolumn{4}{c}{FaceForensics++ \cite{rossler2019faceforensics++}} & \multicolumn{4}{c}{Ted-talks \cite{siarohin2021motion}} \\
    \midrule
    57s & MP & MS & DDIM & Total & MP & MS & DDIM & Total\\
     & 55.6s & 27.5s & 291.7s & 406.8s & 44.7s & 23.6s & 236.2s & 326.6s \\
  \bottomrule
  \end{tabular}
  \caption{Average editing time tested on 500 images per dataset.
  For each image, 57 seconds is needed to finetune a LoRA. The editing phase takes 406.8 seconds per image for the FaceForensics++ dataset \cite{rossler2019faceforensics++} and 326.6 seconds for the Ted-talks dataset \cite{siarohin2021motion}. MP means Motion Prediction time while MS and DDIM represent Motion Supervision time and DDIM Sampling time respectively.}
  \label{tab:time}
\end{table*}

\begin{figure*}[!tb]
  \centering
  \includegraphics[width=.85\textwidth]{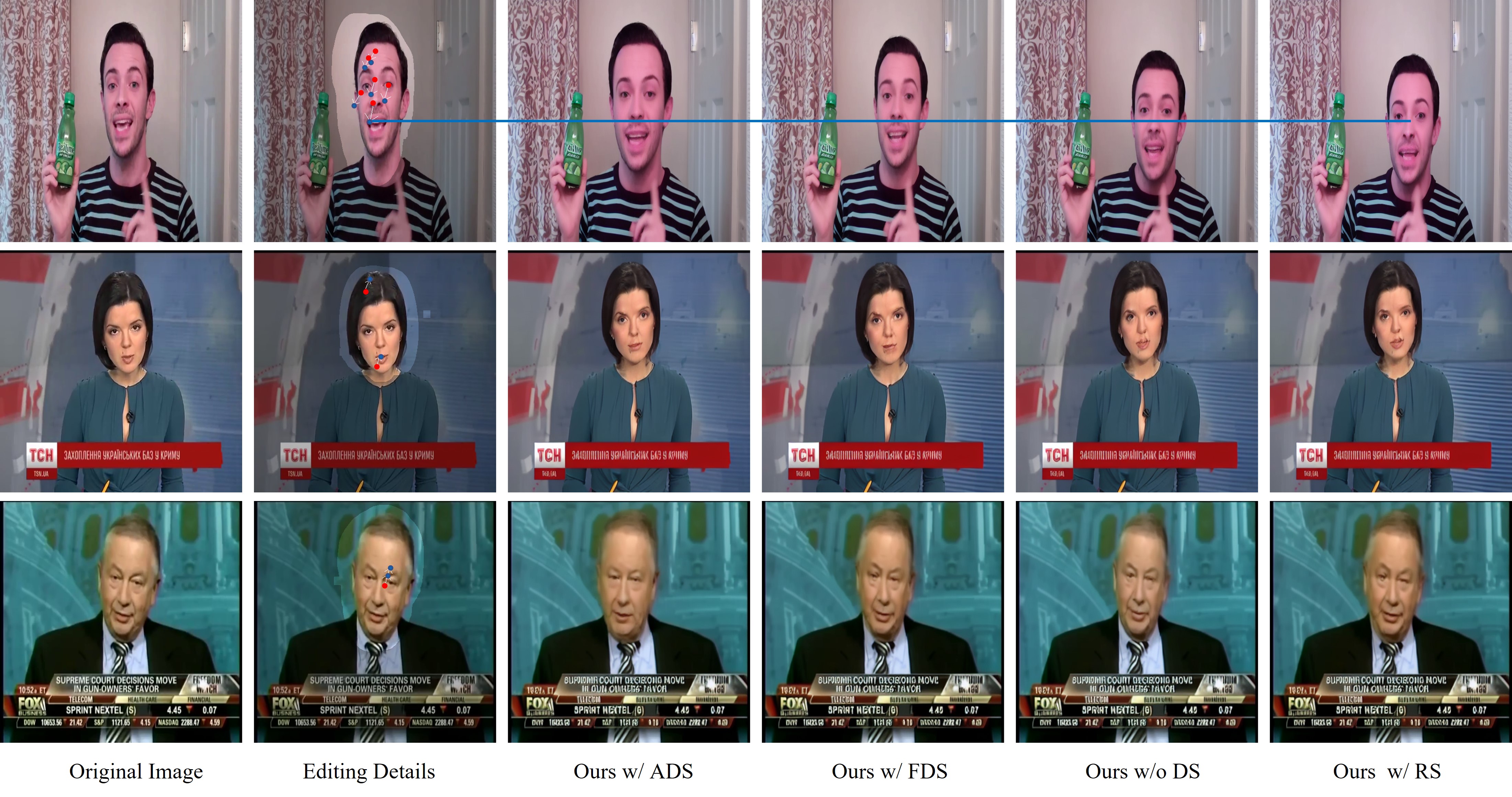}
  \caption{Visual ablation results testing on FaceForensics++ dataset \cite{rossler2019faceforensics++}.
  }
  \label{fig:visual-ablation}
\end{figure*}

\section{Dataset Construction}
\label{sec:dataset_construction}

We use video datasets FaceForensics++ \cite{rossler2019faceforensics++} and Ted-talks Dataset \cite{siarohin2021motion} to construct training set for Motion Prediction module and testing set for evaluating existing methods. FaceForensics++ is a dataset consisting of 1000 original video sequences manipulated with automated face manipulation methods. Every video in FaceForensics++ contains a trackable mostly frontal face without occlusions, which makes it easy to get exactly where the face region is by applying face parsing method to the video. Ted-talks Dataset contains more than 3000 videos that capture a person performing presentation from Internet.

\begin{figure*}[!tb]
  \centering
  \includegraphics[width=.65\textwidth]{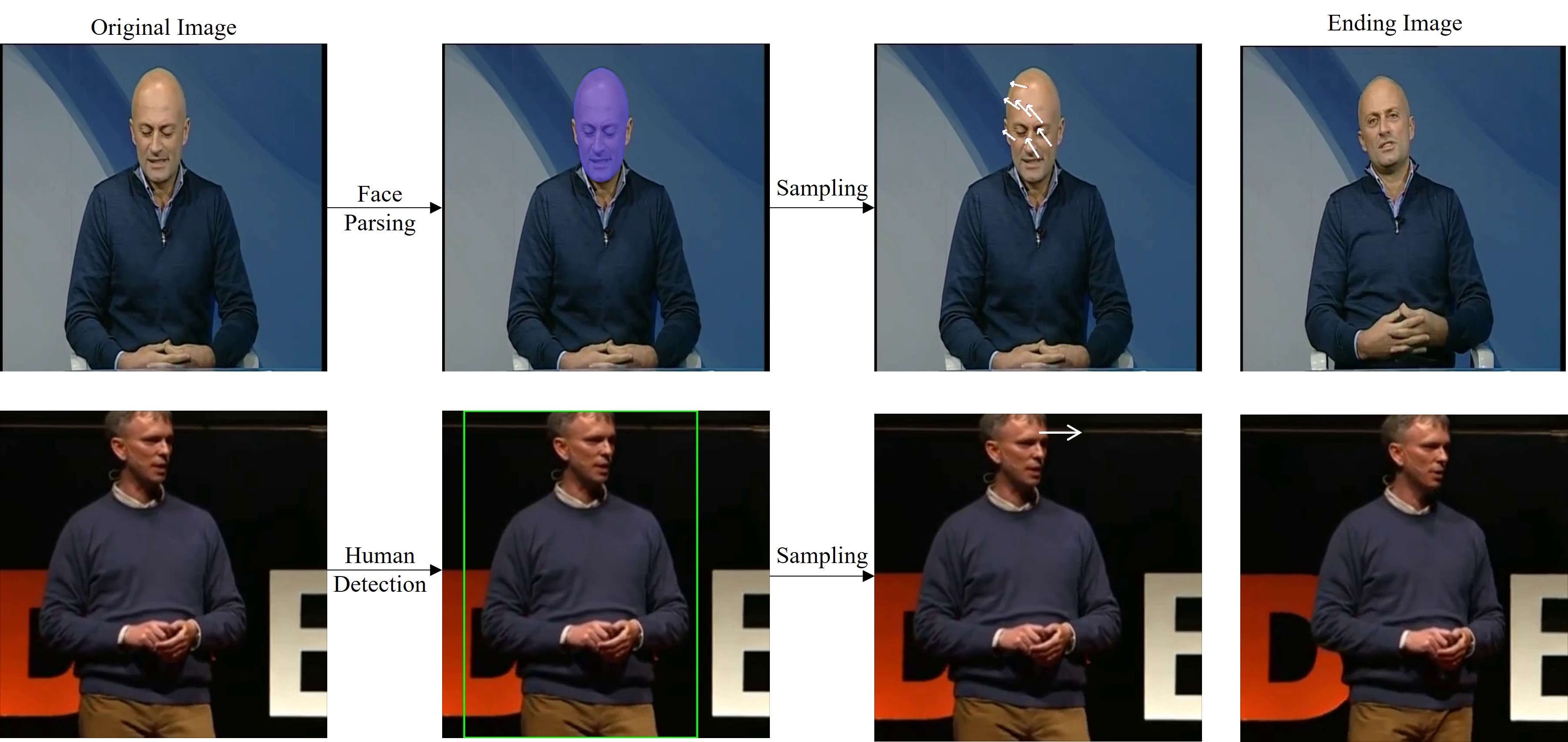}
  \caption{Illustration of dataset construction. We extract editing region with RTTN \cite{lin2021roi} for each video sequence in FaceForensics++ dataset \cite{rossler2019faceforensics++} and YOLOv8 \cite{Jocher_Ultralytics_YOLO_2023} for Ted-talks dataset. Then we sample handle points only in editing region in each frame.
  }
  \label{fig:data_construction}
\end{figure*}

As shown in \cref{fig:data_construction}, we construct the dataset in the following steps:  1) we use Unimatch \cite{xu2023unifying} to predict the optical flow $f \in \mathcal{R}^{S-1 \times H \times W \times 3}$ of the given video $v \in \mathcal{R}^{S \times H \times W \times 3}$. 2) We randomly select a frame $v_{s}$ from the $S$ frames as the starting frame, and then extract editing region with segmentation methods or detection methods.  3) We randomly sample $1 \sim 7$ handle points in the editing region of $v_{s}$ based on the magnitude of its optical flow $f_{s}$ as probability. 4) According to the optical flow predicted by Unimatch, we randomly select a frame in $v_{s+15} \sim v_{s+55}$ as the end frame $v_{e}$, and calculate the ending positions of the handle points in the end frame $v_{e}$ as the target points. 5) The optical flow map $f_{s}$ will be used as groundtruth when training Motion Prediction module. According to the positions of paired handle points and target points, we derive the corresponding delta map $D_{s}$ and heatmap $G_{s}$. As for extracting editing region, for FaceForensics++ dataset, we use the face parsing method RTTN \cite{lin2021roi} to extract the face region as editing region, while for Ted-talks dataset, YOLOv8 \cite{Jocher_Ultralytics_YOLO_2023} is used to detect the human region as editing region. We train two Motion Prediction models using 95\% of the video sequences in FaceForensics++ and Ted-talks trainset respectively, and test them using the rest of video sequences in FaceForensics++ and Ted-talks testset.

\section{More Qualitative Results}
\label{sec:more-quality}
\begin{figure*}[tb]
  \centering
  \includegraphics[width=\textwidth]{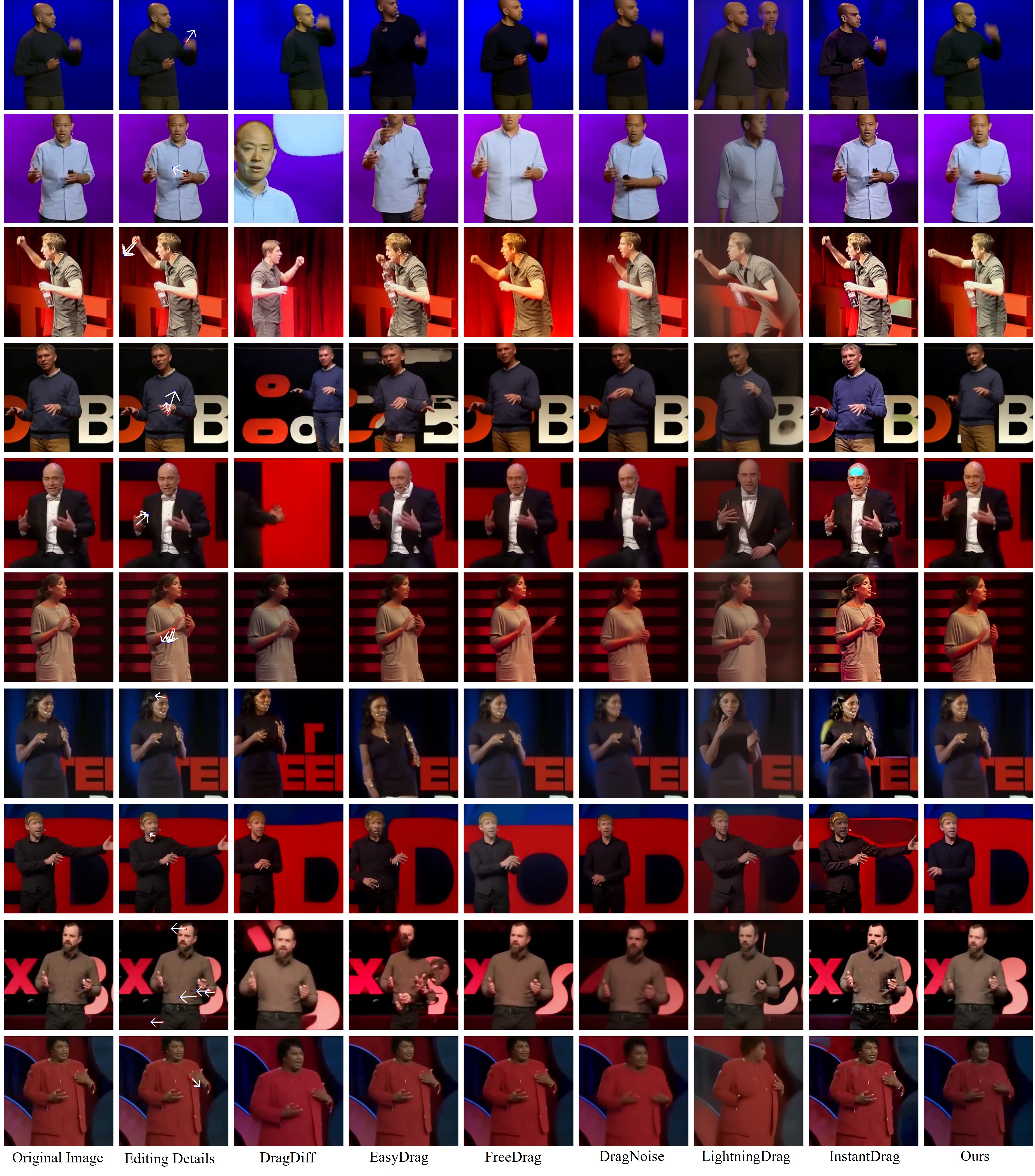}
  \caption{More qualitative results testing on Ted-talks dataset \cite{siarohin2021motion}.
  }
  \label{fig:more_quality_ted}
\end{figure*}

\begin{figure*}[tb]
  \centering
  \includegraphics[width=\textwidth]{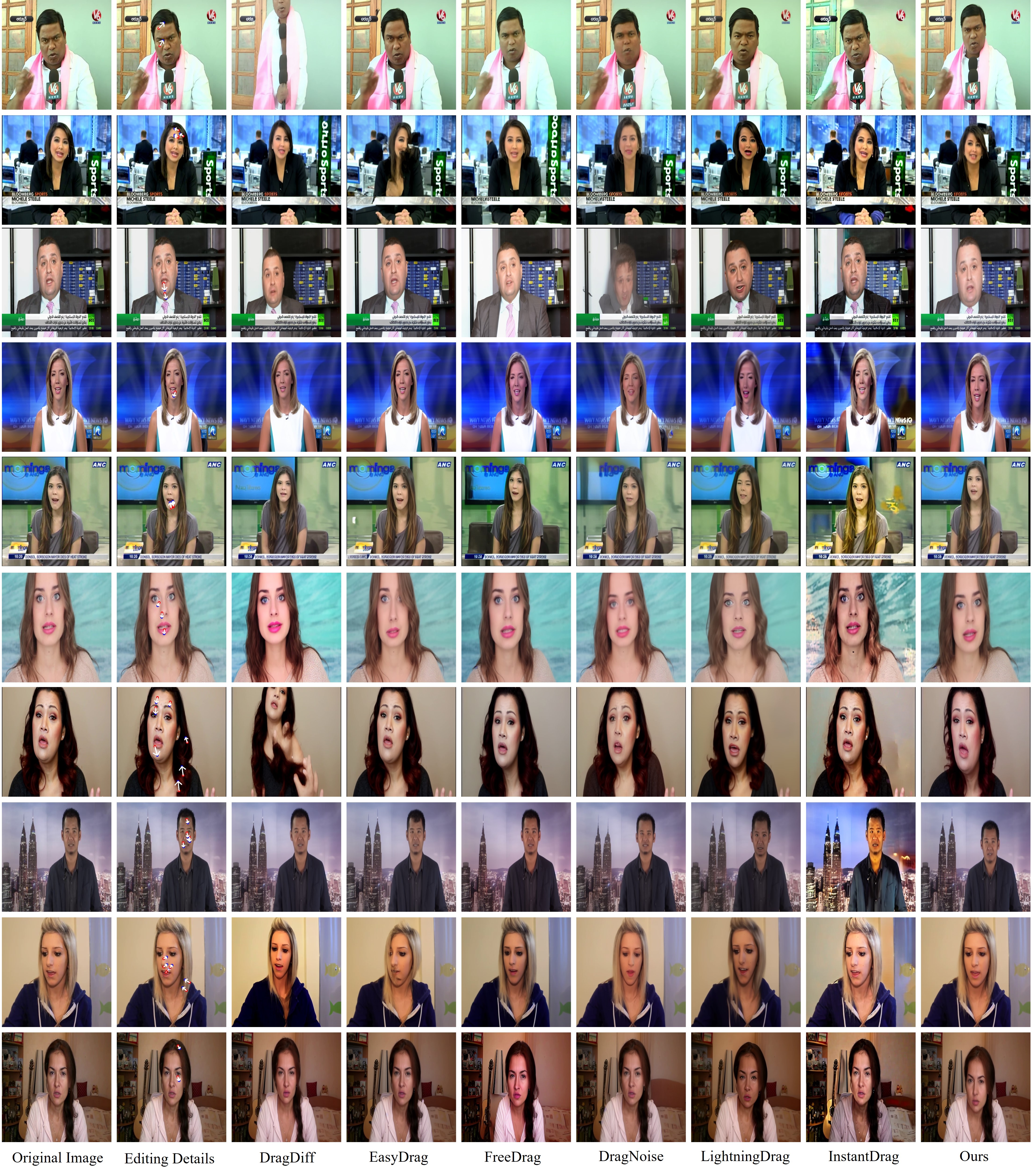}
  \caption{More qualitative results testing on FaceForensics++ dataset \cite{rossler2019faceforensics++}.
  }
  \label{fig:more_quality_ff}
\end{figure*}

In this section, we provide more qualitative comparison between DragDiffusion \cite{shi2023dragdiffusion}, EasyDrag \cite{hou2024easydrag}, FreeDrag \cite{ling2023freedrag}, DragNoise \cite{liu2024drag}, LightningDrag \cite{shi2024instadrag}, InstantDrag \cite{instantdrag} and our proposed DynaDrag. As shown in \cref{fig:more_quality_ted}, our proposed DynaDrag adeptly fulfills editing requirements, including body posture adjustment, body orientation adjustment, and overall body movement in Ted-talks dataset. In the FaceForensics++ dataset \cite{rossler2019faceforensics++} as shown in  \cref{fig:more_quality_ff}, our proposed method effectively modifies facial expressions, facial orientation, and overall facial position. Compared to the previous method, the test results demonstrate enhanced editability and editing accuracy on both datasets. More qualitative results on Ted-talks dataset \cite{siarohin2021motion} and FaceForensics++ dataset \cite{rossler2019faceforensics++} again showcase the superiority of DynaDrag over previous methods.

\begin{figure}[tb]
  \centering
  \includegraphics[width=\linewidth]{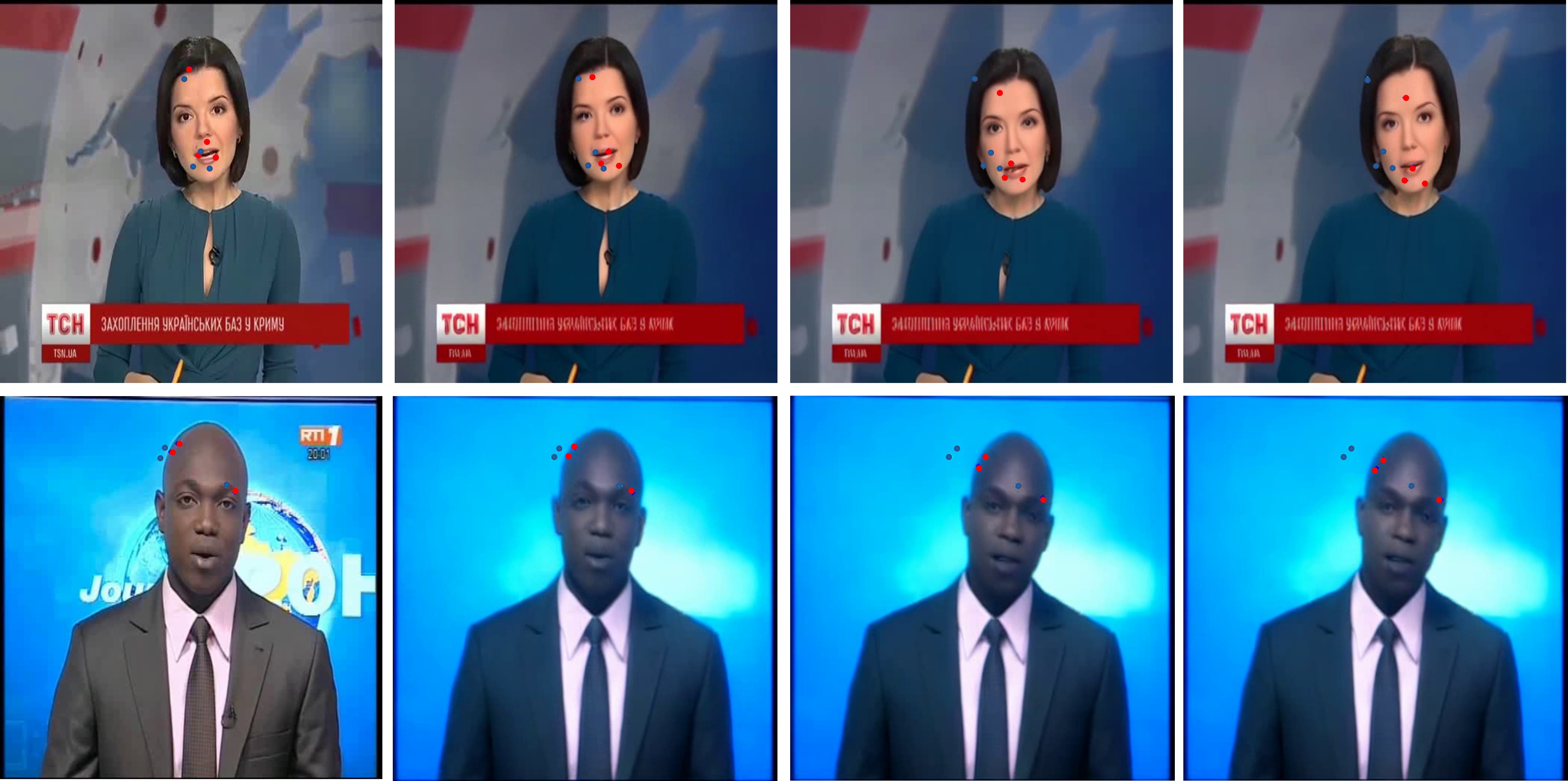}
  \caption{Failure cases in which the handle points (red) should be dragged to the target points (blue). However, Motion Prediction fails to accurately predict the movement of each handle point, leading to undesired editing.
  }
  \label{fig:limitations}
\end{figure}

\section{Visual Ablation on Dynamic Selection}
\label{sec:visual-ablation}
In this section, we present visual illustrations of ablation studies on Dynamic Selection. As depicted in the first subfigure in  \cref{fig:visual-ablation}, the blue line denotes the final position to which the handle point at the nose should be dragged. Notably, only the ADS and FDS have effectively achieved this adjustment. Additionally, upon examination of other pairs of editing points in this subfigure, we observed that during the editing process, the handle point at the nose best reflects the user's editing intention and exhibits the lowest similarity. When DS is not utilized or RS is employed, the points at the nose may undergo excessive dragging due to the influence of other points. However, when DS is activated, ADS can continuously refine the effective points in the edited intermediate results, ultimately producing results that better align with user drag requirements compared to FDS. In the second subfigure in  \cref{fig:visual-ablation}, our approach using ADS and FDS generates fewer artifacts than our approach without DS and our approach with RS. In the third example, our approach using ADS more accurately edits the handle points to the target points.

\section{Limitations}
\label{sec:limitation}
While numerous experiments have demonstrated the superiority of our proposed method over previous drag-style image editing methods \cite{shi2023dragdiffusion,mou2023dragondiffusion}, it is essential to acknowledge that our method also has inherent limitations. By introducing the predict-and-move framework, our method eliminates the need for point tracking, thereby circumventing issues associated with tracking methods. However, the Motion Prediction module introduces its own challenges, notably accuracy issues. In certain instances, the Motion Prediction module fails to accurately predict the movement of each handle point, and in extreme cases, predicts movements contrary to the actual direction, resulting in edited images that do not align with user expectations (as shown in  \cref{fig:limitations}). Furthermore, the training of each Motion Prediction model on a dataset specific to a single scenario restricts the versatility of pre-trained models.

\clearpage
{
    \small
    \bibliographystyle{ieeenat_fullname}
    \bibliography{main}
}

\end{document}